\documentclass{article}

\usepackage{microtype}
\usepackage{graphicx}
\usepackage{booktabs} %

\usepackage{multirow}
\usepackage{multicol}
\usepackage{enumitem}
\usepackage{algorithm2e}
\usepackage{algorithmic}
\usepackage{subcaption}
\usepackage{titling}
\usepackage{comment}
\usepackage{xcolor}
\usepackage{arydshln}

\usepackage{url}

\usepackage{breakurl}
\usepackage{hyperref}
\makeatletter
\renewcommand{\sectionautorefname}{\S\@gobble}
\renewcommand{\subsectionautorefname}{\S\@gobble}
\renewcommand{\subsubsectionautorefname}{\S\@gobble}
\makeatother
\hypersetup{
    colorlinks,
    linkcolor={blue!50!black},
    citecolor={red!50!black},
    urlcolor={blue!80!black}
}

\usepackage{ifthen}

\usepackage{hyperref}

\newboolean{publicversion}
\setboolean{publicversion}{true}

\newcommand{\sysname}{Vidur\xspace}
\newcommand{\sysbench}{Vidur-Bench\xspace}
\newcommand{\syssearch}{Vidur-Search\xspace}

\newcommand{\myx}{$\times$\xspace}
\newcommand{\llamaL}{LLaMA2-70B\xspace}
\newcommand{\llamaS}{LLaMA2-7B\xspace}

\newcommand{\internlmM}{InternLM-20B\xspace}
\newcommand{\qwenL}{Qwen-72B\xspace}
\newcommand{\chat}{LMSys-Chat-1M\xspace}
\newcommand{\arxivL}{Arxiv-Summarization\xspace}
\newcommand{\bwbL}{Bilingual-Web-Book\xspace}
\newcommand{\arxivS}{Arxiv-Summarization-4K\xspace}
\newcommand{\bwbS}{Bilingual-Web-Book-4K\xspace}
\newcommand{\chatshort}{Chat-1M\xspace}
\newcommand{\arxivSshort}{Arxiv-4K\xspace}
\newcommand{\bwbSshort}{BWB-4K\xspace}
\newcommand{\vllm}{vLLM\xspace}
\newcommand{\orcaplus}{Orca+\xspace}
\newcommand{\sarathi}{Sarathi-Serve\xspace}

\newcommand{\vheading}[1]{\vspace{0.05in}\noindent\textbf{#1}}
\newcommand{\sref}[1]{\S\ref{#1}}

\newcommand{\profiler}{\textit{Profiler}\xspace}
\newcommand{\RE}{\textit{Runtime Estimator}\xspace}
\newcommand{\hScheduler}{\textit{Hierarchical Scheduler}\xspace}
\newcommand{\kvcache}{\textit{KV-Cache}\xspace}

\ifthenelse{
\boolean{publicversion}}{
	\newcommand{\grumbler}[3]{}
        \newcommand{\jm}[1]{}
        \newcommand{\ap}[1]{}
        \newcommand{\nk}[1]{}
        \newcommand{\rr}[1]{}
        \newcommand{\amey}[1]{}
        \newcommand{\alexey}[1]{}
        \newcommand{\nitin}[1]{}
}
{%
\newcommand{\grumbler}[3]{\xspace\textcolor{#3}{\bf #1: #2}}
\newcommand{\jm}[1]{\grumbler{Jayashree}{#1}{magenta}}
\newcommand{\ap}[1]{\grumbler{Ashish}{#1}{violet}}
\newcommand{\nk}[1]{\grumbler{Nipun}{#1}{teal}}
\newcommand{\rr}[1]{\grumbler{Ram}{#1}{cyan}}

\newcommand{\amey}[1]{\grumbler{Amey}{#1}{teal}}
\newcommand{\alexey}[1]{\grumbler{Alexey}{#1}{teal}}
\newcommand{\alexey}[1]{\grumbler{Nitin}{#1}{orange}}

}

\setlist[itemize]{noitemsep, topsep=0pt}

\usepackage[accepted]{mlsys2024}

\mlsystitlerunning{\sysname: A Large-Scale Simulation Framework for LLM Inference}

\begin{document}

\twocolumn[
\mlsystitle{\sysname: A Large-Scale Simulation Framework for LLM Inference}

\begin{mlsysauthorlist}
\mlsysauthor{Amey Agrawal}{gt,int}
\mlsysauthor{Nitin Kedia}{msr}
\mlsysauthor{Jayashree Mohan}{msr}
\mlsysauthor{Ashish Panwar}{msr} 
\mlsysauthor{Nipun Kwatra}{msr} \\
\mlsysauthor{Bhargav S. Gulavani}{msr}
\mlsysauthor{Ramachandran Ramjee}{msr}
\mlsysauthor{Alexey Tumanov}{gt}
\end{mlsysauthorlist}

\mlsysaffiliation{gt}{Georgia Institute of Technology, USA.}
\mlsysaffiliation{msr}{Microsoft Research India}
\mlsysaffiliation{int}{Part of work done as an intern at Microsoft Research India.}

\mlsyscorrespondingauthor{Amey Agrawal}{ameyagrawal@gatech.edu}

\mlsyskeywords{Machine Learning, MLSys}

\vskip 0.3in

\begin{abstract}

Optimizing the deployment of Large language models (LLMs) is expensive today since it requires experimentally running an application workload against an LLM implementation while exploring large configuration space formed by system knobs such as parallelization strategies, batching techniques, and scheduling policies.
To address this challenge, we present \sysname{} -- a large-scale, high-fidelity, easily-extensible simulation framework for LLM inference performance.
\sysname models the performance of LLM operators using a combination of experimental profiling and predictive modeling, and evaluates the end-to-end inference performance for different workloads by estimating several metrics of interest such as latency and throughput.  We validate the fidelity of \sysname on several LLMs and show that it estimates inference latency with less than 9\% error across the range. 
Further, we present \syssearch, a configuration search tool that helps optimize LLM deployment. \syssearch{} uses \sysname{} to automatically identify the most cost-effective deployment configuration that meets application performance constraints. For example, 
\syssearch finds the best deployment configuration for \llamaL in one hour on a CPU machine, in contrast to a deployment-based exploration which would require 42K GPU hours -- costing 218K dollars. 
Source code for \sysname is available at \url{https://github.com/microsoft/vidur}.
\end{abstract}

]

\printAffiliationsAndNotice{}  %

\section{Introduction}
\label{sec-intro}

Large language models (LLMs) can learn from and generate natural language texts on a massive scale. LLMs such as GPT-3/4~\cite{gpt3-brown2020language, sparksofagi}, LLaMA~\cite{llamaarxiv}, and Phi~\cite{li2023textbooks} have demonstrated impressive performance on various natural language processing (NLP) tasks. However, LLM inference -- the process of using an LLM to produce natural language outputs based on some input -- is expensive. For example, the cost of serving ChatGPT is estimated to be $\$694$K per day~\cite{semi}. 

An LLM inference provider faces several challenges in optimizing LLM deployment. 
First, the provider has to choose a model parallelization strategy such as the number of tensor parallel dimensions, number of pipeline stages, number of replicas, etc. Second, the operator has to choose between different scheduling algorithms (e.g., Orca~\cite{orca}, vLLM~\cite{vllmpaper}, Sarathi-Serve~\cite{sarathi-serve}). 
Third, the provider has to determine several configuration parameters, such as maximum batch size (BS),  wait time for batching, as well as algorithm specific parameters (e.g., chunk size in Sarathi, watermark fraction in vLLM) to satisfy the desired throughput and latency constraints. 
Finally, they have to generate representative workload traffic to test out each of their models on an experimental testbed with each of the different combinations above. {\it Systematically optimizing deployment of tens of models with hundreds of configuration options is expensive and impractical.}

\begin{figure*}[t]
    \centering
    \begin{subfigure}{0.55\linewidth}
        \includegraphics[scale=0.37]{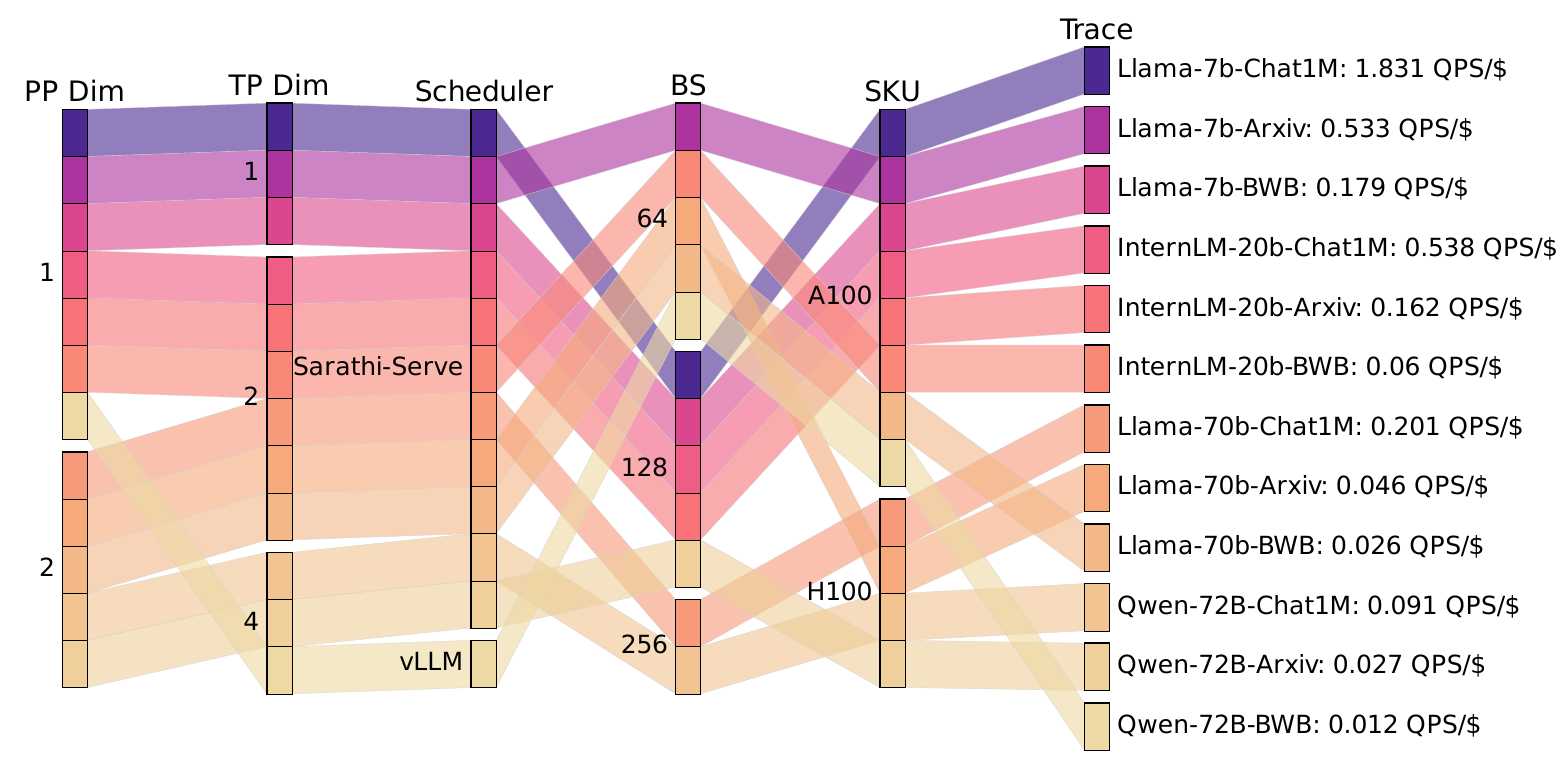}
        \caption{\small
            \textbf{Optimal configurations:} Color bands correspond to the optimal config for each of the 12 model-trace pairs with corresponding throughput achieved per dollar. %
            }
        \label{fig:intro:optimal-configs}
    \end{subfigure}
    \hfill
    \begin{subfigure}{0.40\linewidth}
        \centering
        \includegraphics[scale=0.50]{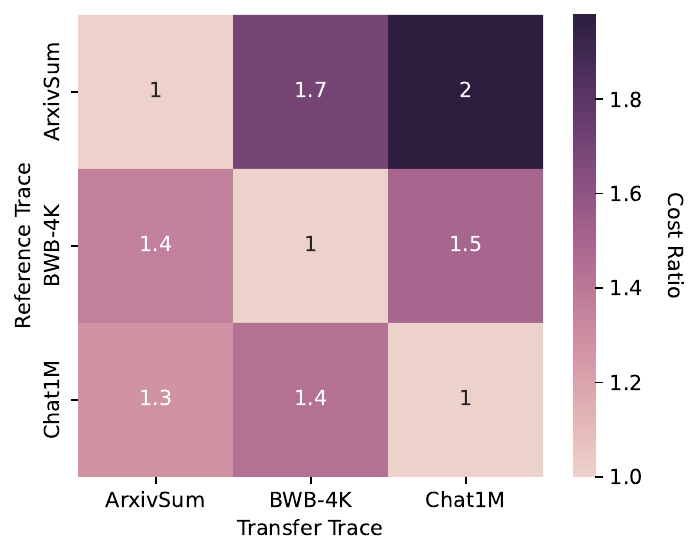}
        \caption{\small
            \textbf{Cost of mis-configuration}: the optimal config on one trace used for another results in up to 2\myx cost difference (\llamaL).
        }
        \label{fig:intro:cost-of-misconfig}
    \end{subfigure}
    \caption{\small
        \textbf{Both the model and workload matter for the optimal deployment configuration}. Optimal configurations for each model-trace pair is shown in (a). Throughput/cost can differ significantly for the same model if the workload is changed as shown in (b).
    }
    \label{fig:intro:best-cofig}
\end{figure*}

This cost is further exacerbated by our observation that optimal configuration is a function of a model-trace pair, i.e., optimal configuration also depends on application workload characteristics (\autoref{fig:intro:optimal-configs}). In fact, an optimal config obtained on one trace could be sub-optimal by a factor of up to 2\myx (\autoref{fig:intro:cost-of-misconfig}) when applied to the same model on a different trace. With both new models and new traces being released almost daily, the cost of identifying the optimal deployment configuration becomes prohibitively expensive.

To tackle this challenge, we present \sysname{} -- a large-scale, high-fidelity and extensible LLM inference performance simulator, and \syssearch{} -- a configuration search tool. Together, they enable \textit{fast} and \textit{inexpensive} exploration of LLM inference performance under a variety of deployment scenarios.

Simulating LLM inference poses several unique challenges that are not addressed in prior work that simulate the performance of deep neural network (DNN) training~\cite{daydream, habitat, recommender_modeling}. 
First, LLM inference predictions have to be accurate at much finer time granularity  compared to training jobs where each iteration runs for hundreds of milliseconds. Second, unlike training where batch sizes are typically fixed, the input sizes during inference can vary drastically. The difference in input sizes stems from varying sequences lengths of different requests, as well as the interleaving of prefill and decode stages depending on the scheduling strategy, resulting in significant variations in iteration latency. Since it is infeasible to experimentally profile the performance of the model for all possible input sizes,  the simulator has to rely on a mixture of careful profiling and a prediction strategy for unprofiled input sizes.  Third, small errors in predictions lead to cascading effect due to the dynamic and stateful nature of inference workloads, thus inference simulators need to provide extremely accurate per-iteration predictions to get good fidelity at high request arrival rates.

\vheading{\sysname}. To address these challenges, \sysname{} uses the key insight that the large majority of LLMs share similar architectures that can be decomposed into a small set of {\it token-level, sequence-level and communication operators}. Thus, \sysname{} takes in a model specification and first identifies various operators and a minimal set of input sizes that need to be profiled experimentally. \sysname then builds a fine-grained runtime estimator that accurately predicts kernel performance on input sizes that might not have been profiled. Using the estimator, \sysname takes a specification of deployment configuration and workload, and predicts a variety of request-level metrics such as Time to First Token (TTFT), Time Between Tokens (TBT), latency, throughput, as well as cluster-level metrics such as Model Flops Utilization (MFU) and memory utilization.

We demonstrate the fidelity of \sysname across a range of models, hardware and cluster configurations. \sysname accurately predicts request-level LLM inference performance with under 9\% error rate, and mimics overall cluster metrics for large-scale workloads and traces with high fidelity.

\vheading{\sysbench}. We find that the workload has a considerable impact on output metrics of interest in LLM inference. For example, variations in the number of input tokens, number of decode tokens and batch size can impact performance dramatically~\cite{sarathi}. We observe that there is no standardized benchmark suite available today to comprehensively evaluate LLM inference performance. Thus, we introduce \sysbench to address this gap. \sysbench is an easily extensible collection of workload traces along with several existing batching and scheduling policies such as vLLM~\cite{vllmpaper}, Orca ~\cite{orca}, FasterTransformer~\cite{fastertransformer} and Sarathi-Serve~\cite{sarathi-serve}.

\vheading{\syssearch}. Finally, we present \syssearch{} to help LLM inference providers optimize their deployment. \syssearch{} uses \sysname{} to automatically search over hundreds of deployment configurations to identify the highest throughput/cost configuration for a given model, workload pair. For example, for \llamaL,  across a pool of A100 / H100 GPUs, \syssearch{} is able to identify the best configuration about one hour on a 96-core CPU cores that costs \$9.93 per hour on Microsoft Azure, as opposed to an actual deployment-based exploration that would have taken 42K GPU hours, costing approximately \$218K.

In summary, this paper makes the following contributions.
\begin{itemize}
    \itemsep0em 
    \item  \sysname: an LLM inference simulator that predicts key performance metrics of interest with high-fidelity (\sref{sec-design})
    \item  \sysbench: a benchmark suite comprising of various workload patterns, schedulers and serving frameworks, along with profiling information for popular hardware like A100 and H100 GPUs (\sref{sec-benchmark}).
    \item \syssearch: a configuration search tool that helps optimize deployment by identifying the highest throughput per dollar configuration (\sref{sec-syssearch}).
\end{itemize}

\section{Background and Motivation}
\label{sec-bgk}
\label{sec:back}

\subsection{Overview of LLMs}
\label{sec:back:overview}
LLMs utilize the transformer architecture based on the self-attention mechanism~\cite{paper:attention} as their core building block. The self-attention mechanism helps a language model learn the relationship between different elements of an input sequence and subsequently produce the output sequence. An LLM consists of two dominant sub-modules, self-attention and multilayer perceptron (MLP). Various LLMs have been developed in recent years using a variation of these modules (e.g., GPTs, LLaMAs, Falcons). Primarily, these models differ only in terms of the embedding size, the number of transformer blocks, and the attention mechanism used by the model.

\subsection{LLM Inference Efficiency Optimizations}
\label{sec:back:efficiency}
LLM inference request processing consists of two distinct phases -- prefill and decode. The prefill phase processes the entire user input prompt and produces the first output token. Subsequently, output tokens are generated one at a time in an autoregressive manner. During this decode phase, the token generated in the previous step is passed through the model to generate the next token until a special \textit{end-of-sequence} token is generated at which point the request processing completes. The decode process requires access to the key and value activations of the previously processed tokens to perform the attention operation. To avoid repeated computation, contemporary LLM inference systems store them in \kvcache.

Given the immense cost of LLM inference, LLM inference efficiency has become an active area of systems research. To this end, multiple optimization mechanisms have been proposed recently. Each of these techniques make different tradeoffs. For cost effective inference, right set of optimizations should be used be composed based on the specific application requirements. For example, Tensor Parallelism (TP) is a common strategy to parallelize LLM inference~\cite{megatron,efficiently-scaling-transformer-inference}. 
TP shards each layer across the participating GPUs by splitting the model weights and \kvcache equally across GPU workers. TP 
(1) improves inference throughput with higher batch sizes, 
(2) lowers the latency of inference by splitting each operator across multiple GPUs. However, TP involves frequent blocking communication between workers, and thus requires expensive hardware with specialized high bandwidth interconnects like NVLINK. Alternatively, Pipeline Parallelism (PP) is another parallelization strategy in which the model is partitioned into stages of consecutive transformer blocks. Each GPU is responsible for computing a stage and output activations are transferred across GPU boundaries via send/recv operations. PP has a much more favorable compute-communication ratio compared to TP, but can suffer from pipeline bubbles (stalls due to imbalance between stages). 

Recently, \citealt{sarathi-serve} identified an inherent tradeoff in LLM inference scheduler design and proposed classification of existing LLM inference schedulers into two categories -- prefill prioritizing \cite{orca, vllmpaper} and decode prioritizing \cite{fastertransformer}. Prefill prioritizing schedules achieve higher throughput, by generating schedules with higher batch sizes, but suffer higher latency cost. Decode prioritizing schedulers can achieve low latency but at the cost of lower throughput \cite{vllmpaper}. Sarathi-Serve \cite{sarathi-serve} tries to mitigate this tradeoff by utilizing the computational slack in decode phase. Another set of recent works, Splitwise \cite{splitwise} and DistServe \cite{distserve} tackle this latency-throughput tradeoff by splitting the computation of prefill and decodes on separate devices.

\noindent{\bf Takeaway:} \textit{Various systems optimizations provide a rich cost-latency tradoff. The right techniques to use depend on the application requirements and hardware availability.} 

\subsection{LLM Inference Configuration Space}
Control knobs like parallelism strategy, choice of scheduler, chunk size, batch size, SKU, etc. induce a large configuration space (\autoref{fig:intro:optimal-configs}) for LLM deployment. Furthermore,
we make an important observation (\autoref{fig:intro:best-cofig}) that the optimal configuration (defined as a combination of specific choices for each control knob) is not simply a function of a specific model. But rather, the optimal configuration varies as a function of both the model $m$ and the trace $t$ evaluated on that model. Thus the complexity of configuration search is $O(|M|\cdot |T|)$, where $M$ is a set of all models of interest and $T$ is a set of workloads. With a rapid increase in both the number of models and downstream applications, the cost of optimal configuration search simply doesn't scale. And yet, misconfiguration is prohibitively expensive. For example, ~\autoref{fig:intro:cost-of-misconfig} shows that using the optimal configuration of one trace can have up to 2\myx
cost differential on a different trace.

\noindent{\bf Takeaway:} \textit{There is no single best deployment configuration for a model -- rather the choice of configuration should be made in a workload-aware fashion.}

With the cost of obtaining a single point in~\autoref{fig:intro:best-cofig} as high as \$97k,  the high cost of misconfiguration, and the size of the search space growing with both models and traces, this begs a fundamental research question: \textit{is it possible to find a performant configuration without requiring access to expensive experimental resources at a fraction of the cost?}
We explore this question in depth by proposing a simulation-based approach for LLM configuration search with \sysname, reducing the cost by several orders of magnitude.

\section{Challenges in Simulating LLM Inference}
State-of-the-art DNN simulation frameworks (Daydream~\cite{daydream}, Habitat~\cite{habitat} and Proteus~\cite{proteus}) focus on training jobs. Building a large-scale inference simulator, especially for LLMs, involves multiple challenges that are not addressed by the existing simulators. We enumerate them in detail below.

\vheading{Time Scale.} Conventional DNN training workloads are typically compute-bound workload where each iteration  executes for 100s of milliseconds~\cite{daydream}. In comparison, LLM inference is a far more latency-sensitive task where iterations can be much shorter (a few milliseconds each)~\cite{orca, vllmpaper}. Therefore, simulating LLM inference requires predicting iteration times at a much finer granularity. %

\vheading{Varying Iteration Times.} Compared to traditional DL workloads where each iteration performs the same amount of compute and has predictable minibatch latency~\cite{gandiva}, latency of different iterations can vary significantly during LLM inference. The variation in inference runtimes come from multiple sources. First, LLM inference consists of different phases -- prefill and decode, each with a different compute characteristic and runtime. Second, the requests being processed may have a large variation in their sequence length (due to varying prompt lengths or number of decode tokens generated), resulting in varying runtimes. Third, the batch size during online inference keeps varying depending on the system load and workload characteristics. Moreover, the composition of a batch can accommodate requests from both prefill and/or decode phases, again adding to the runtime variation.

\vheading{Cascading Errors.} In training workloads, the batch composition is uniform across all batches, and the execution of each batch is independent. However, during inference, requests arrive in the system dynamically, and if the runtime prediction of any batch has significant errors, that can change in the batching pattern. Thus small errors in individual batch predictions cascade over time and lead to aggregate errors.

\section{\sysname}
\label{sec-design}

\begin{figure}[t]
    \centering
    \hspace{-1.15em}
    \includegraphics[width=0.47\textwidth]{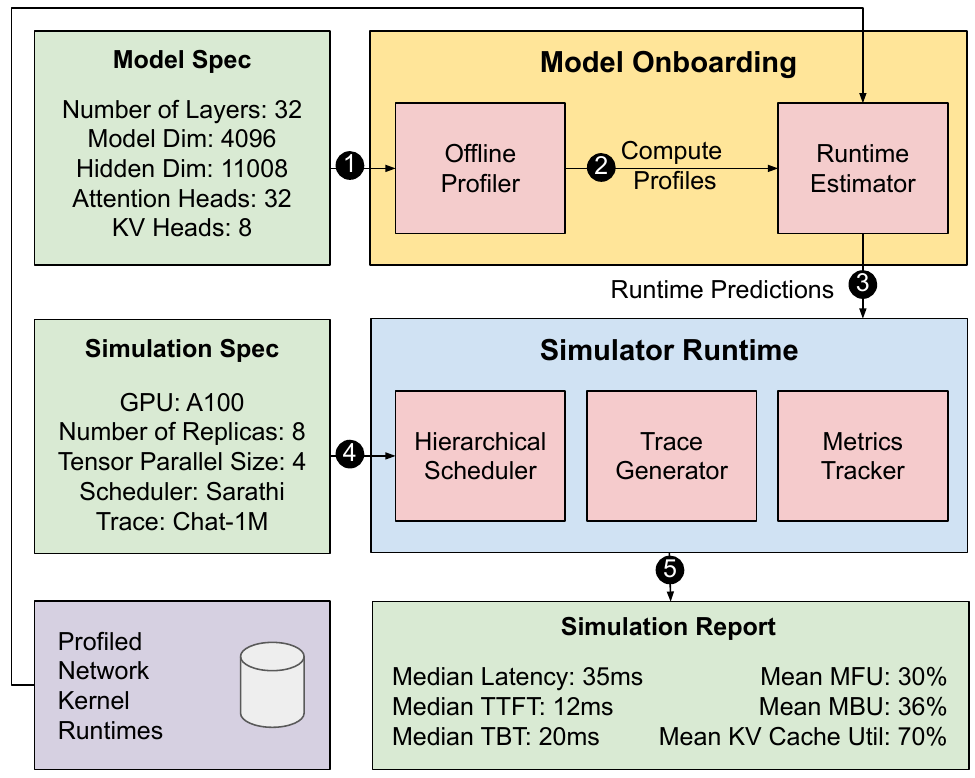}
    \caption{\small
        \sysname Simulator High Level Architecture.}
    \label{fig:hld}
\end{figure}

\sysname leverages domain knowledge to provide high-fidelity performance estimations of LLM inference. It emulates the behavior of all layers of the inference stack, including both the model execution and the various tiers of request scheduling, at both replica as well as the cluster level.

\subsection{Key Insights}
\vheading{LLMs Share Key Architectural Properties.} The large majority of LLMs share fundamentally similar architectures with small differences in the choice of activation functions, normalization layers, residual connections, etc. This allows us to use a common declarative model specification format that captures the essential architectural choices of various models. Another consequence of this architectural uniformity is that \sysname only needs to model a small number of compute operators that are shared across all model families.

\vheading{Operation Triaging for Runtime Prediction.} In a running batch, each request may be associated with varying numbers of \kvcache and \textit{query} tokens, leading to a vast combinatorial input space. Consequently, profiling every possible combination to predict operation runtimes is not feasible. Instead, we observe that LLM operators can be classified into different categories. For instance, execution time of some operations depend on the total context length of all the requests in the batch whereas for others, it depends only on the number of tokens in the current iteration. This classification allows us to design tailored runtime prediction strategies for each operator type.

For example, we observe that apart from the attention kernel, all other operations are independent of request history. During the decode phase, the MLP layer would take the same amount of compute irrespective of the number of input or output tokens processed previously. Profiling the attention kernel requires modeling history of each request. However, since the attention operation during decode is largely a memory-bound operation \cite{flashattention, sarathi}, we find that it is sufficient to model the total amount of \kvcache to be fetched in a batch of requests to determine the kernel runtime (\sref{sec:profiler}).

\vheading{Automatic Profiling for Parallelism Strategies.} 
Each model parallel configuration has different memory, compute, and network communication characteristics. A naive profile and replay approach would require a separate profiling run for each parallelism configuration, which can be expensive. In contrast, \sysname incorporates the domain knowledge about LLM parallelism strategies, which allows it to identify the subset of computation that is performed on each device. During the profiling phase, we automatically identify the tensor sharding configurations for each operator from a declarative specification of the model. Consequently, \sysname can simulate various parallelization schemes with minimal profiling performed on a single GPU.

\subsection{System Overview}

\sysname primarily has two phases of processing. First is the model onboarding phase wherein the model specification is used to generate a set of compute operators to be profiled. The \sysname profiler (\sref{sec:profiler}) collects the runtime characteristics for the identified operators and feeds them to the runtime estimator. To minimize the cost barrier of adding new models to the system, we collect minimal data during the profiling phase and then train small machine-learning models to generate predictions over a large range of parameters that these operation could be triggered on during simulation. This phase is handled by \sysname's runtime estimator (\sref{sec:re}), which produces operation-wise runtime lookup tables that can be later used during simulation.

Once the model is onboarded, the user can perform simulations using various scheduling policies, and parallelism strategies, across a wide range of workloads supported by \sysbench (\sref{sec-benchmark}). At the core of our event-driven simulator is a pluggable \hScheduler (\sref{sec:scheduler}), which supports several popular batching strategies alongside memory planning and management capabilities. The simulator provides detailed metrics that capture both the request (normalized latency, time-to-first-token, time-between-tokens, etc.) and cluster (Model FLOPs utilization, \kvcache utilization, etc.) performance metrics. The end-to-end process flow in \sysname is illustrated in~\autoref{fig:hld}. %

\subsection{\profiler}
\label{sec:profiler}

To efficiently profile the runtime characteristics of LLMs, we leverage the insight that the large majority of LLMs share fundamentally similar architectures with small differences in the choice of activation functions, normalization layers, residual connections, etc. %

\vheading{Operator Triaging.} The profiler analyzes different operators to identify their input dependencies. We find that all the operators can be placed on one of the three buckets:

\begin{itemize}
    \item \textit{Token-level Operators:} The operand dimensions for operations like linear, and activation functions depend on model architecture, however, their runtime only depends on the total number of tokens being processed (prefill plus decode) in the batch.
    \item \textit{Sequence-level Operators:} The attention operation depends not only on the number of tokens in the current batch but also the context length of each request.
    \item \textit{Communication Operators:} The runtime of communication operations like \textit{all-reduce} and \textit{all-gather} depend only on the amount of data to be transferred, independently of the model architecture.
\end{itemize}

\vheading{Profiling Token-level Operators.} There are two broad categories of token-level operators - matrix multiplications and simple point-wise apply or reduction operations, like addition, normalization, and activation functions. Based on the model specification, we generate all the different tensor parallel sharding configurations and profile each combination. This approach allows us to obtain traces for different parallelism configurations while profiling on a single GPU. We use standard PyTorch kernels for profiling these operations and measure their performance using CUPTI \cite{cupti}.

\vheading{Profiling Sequence-level Operators.} \label{heading:profiling-seq-level-ops} Batching sequence-level operators such as the attention kernels is sensitive to the context length of the requests in the batch, thereby exploding the state space of inputs  to profile.%
We use several techniques to address this problem. First, we separately profile the attention kernels for prefill and decode phases due to their difference in compute characteristics. 

While processing the prefill attention, we observe that the attention time for each prefill is quadratic in its length. Suppose we have a batch of $P$ prefills of length $p_i$, where $i$ varies from $1$ to $P$. The cost of prefill attention for the whole batch is therefore proportional to \(\Sigma_{i=1}^P p_i^2\). To approximate the runtime of this batch we predict the runtime of an \textit{equivalent} batch of a single prefill of length \(\sqrt{\Sigma_{i=1}^P p_i^2}\).

In contrast to prefill, we notice that the attention decode operation is largely memory-bound~\cite{flashattention, sarathi}. %
As a result, the runtime of this operation is mainly determined by the total data volume that needs to be fetched from the \kvcache and not the exact split of context lengths between different requests in the batch. In practice, the attention kernel might not be able to effectively parallelize \kvcache fetch operation when there is a large skew between the context length of different requests in a batch. However, we observe that sequence parallel attention kernels such as PagedAttention v2 \cite{vllmpaper}, and FlashDecoding \cite{flashdecoding} can effectively handle such skews, and thus it is sufficient to model decode based on total \kvcache reads. %

\vheading{Profiling Communication Operators.}  There are three collective operations that are frequently used in LLM inference, namely, \textit{all-reduce}, \textit{all-gather} (used for tensor parallelism) and \textit{send-recv} (used for pipeline parallelism). %
Since these operations don't depend on model-specific characteristics, we independently profile these kernels ahead of time in a model-agnostic manner for different topologies.

\subsection{\RE}
\label{sec:re}

Collecting profiling data for every possible input combination across all the operators is prohibitively expensive. Therefore, we collect a limited set of data points and rely on small machine-learning models to interpolate the runtimes. \RE first trains these models using the profiled data, and then generates runtime estimates for a large range of input tensor dimensions which it encounters in end-to-end simulation.

Prior DL training simulators \cite{habitat,recommender_modeling} train Multi-layer Perceptron (MLP) models for opaque operations like matrix multiplications which are provided by closed-source third-party libraries like CUBLAS \cite{cublas} and cuDNN \cite{cudnn}. However, training MLPs requires a large amount of data and results. On the other hand, simple polynomial regression does not capture the non-linear runtime characteristics of CUDA kernels due to phenomenons like tile and wave quantization~\cite{tilewavequantization}. For our scenario, we find that random forest (RF) regression models achieve the right balance between data frugality and fidelity. %

\subsection{\hScheduler}
\label{sec:scheduler}

In \sysname we adopt a three-tier hierarchical scheduler architecture, that provides a powerful and extensible interface. First is the global scheduler, that is responsible for request routing in \sysname. In addition to standard load balancing policies like round-robin and least outstanding requests, we also support stateful scheduling policies, where routing decisions can be deferred to a later point in time, which can be helpful under busty workloads where early binding routing decisions can hurt performance. 

Second is the replica scheduler that encapsulates two key responsibilities; batching and memory management. The replica scheduler contains a memory planner, which uses the model specification and parallelism configuration to compute the memory available for \kvcache. This information is then used by the memory manager to provide high-level management APIs that are used to implement custom batching policies. \sysname currently supports five batching policies, FasterTransformers \cite{fastertransformer}, Orca \cite{orca}, Sarathi-Serve \cite{sarathi-serve}, vLLM \cite{vllmpaper} and LightLLM \cite{lightllm}. The high-level API support provided by \sysname makes it extremely simple to implement new batching policies; all the aforementioned policies have been implemented each in less than 150 lines of Python code in our simulator %

The final component of our scheduling stack is the replica stage scheduler, which handles the scheduling of microbatches within a pipeline stage. While we currently only support synchronous pipeline parallel scheduling policy, in the future, we aim to extend the replica stage scheduler to emulate various optimizations like asynchronous communication, sequence parallelism \cite{terapipe} and speculative pipelined decoding \cite{speed}. %

\section{\sysbench}
\label{sec-benchmark}
\sysbench is a benchmark suite for easy evaluation performance evaluation of LLM inference systems that comprises of plug-and-play support for a variety of (a) workload patterns, (b) scheduling, batching, and routing policies, and (c) serving frameworks.

\begin{table*}[]
\center
\scalebox{0.65}{
\begin{tabular}{c|c|c|cll|cll|cll} \hline
\multirow{2}{*}{Dataset} & \multirow{2}{*}{Content} & \multirow{2}{*}{\# queries} & \multicolumn{3}{|c}{\# prefill tokens} & \multicolumn{3}{|c|}{\# decode tokens} & \multicolumn{2}{c}{P:D Ratio} \\ \cline{4-11}
 & &  & mean & median & p90 & mean & median & p90 & median & std dev\\ \hline
\chat[~\citealt{lmsyschat1m}] & Natural language conversations & 2M & 786 & 417 & 1678 & 215 & 141 & 491 & 2.3 & 236 \\
(\chatshort) & \chat with max 4k total tokens & 2M & 686 & 417 & 1678 & 197 & 139 & 484 & 2.3 & 228 \\
\arxivL[~\citealt{cohan-etal-2018-discourse}] & Summarization of arxiv papers & 203k & 9882 & 7827 & 18549 & 411 & 228 & 475 & 35.4 & 81 \\
(\arxivSshort) & \arxivL with max 4k total tokens & 28k & 2588 & 2730 & 3702 & 291 & 167 & 372 & 15.7 & 16 \\
\bwbL[~\citealt{jiang-etal-2023-discourse}] & Document-level English--Chinese parallel dataset & 195k & 2418 & 2396 & 3441 & 3654 & 3589 & 5090 & 0.66 & 0.23 \\
(\bwbSshort) & \bwbL with max 4k total tokens & 33k & 1067 & 1037 & 1453 & 1612 & 1601 & 2149 & 0.65 & 0.37 \\
\hline
\end{tabular}
}
\caption{Details of the workloads curated from open-source datasets.}
\label{table:workloads}
\end{table*}

\subsection{Datasets and workloads}
The overall performance of LLM inference is highly sensitive to the type of workloads such as the number of input and output tokens in a given query e.g., the decode phase can be as high as $200\times$ more expensive than the prefill phase~\cite{sarathi}. Different workload patterns can therefore influence system performance in complex ways. For instance, vLLM incrementally allocates physical memory for the \kvcache in order to fit a large batch size on the GPU. This works well when the number of decode tokens is high e.g., in chat applications~\cite{lmsyschat1m}. In contrast, incremental memory allocation is less useful if the prompt length is much higher than the number of output tokens as in summarization tasks.

\sysbench provides a set of workloads curated from publicly available datasets (see \autoref{table:workloads}). These can be used to evaluate system performance for varying request types, arrival rates etc. or to tune the performance sensitive parameters of various components in the serving system. 

\subsection{Performance metrics}
\sysbench provides a comprehensive set of system-level performance metrics as discussed below:

\noindent\textbf{Operator-level metrics.} This includes each operator's input size and execution time which can be used to identify and optimize the heavy-duty operators eg. \textit{attn\_prefill}, \textit{mlp\_up\_proj} etc. \\
\noindent\textbf{Request-level metrics.} These include per-request metrics such as the scheduling delay, prefill completion time, time-to-first-token (TTFT), and time-between-tokens (TBT). Furthermore, any additional metrics of interest can be easily added, e.g., we added support to track how many times vLLM preempts or restarts each request when it runs out of GPU memory for \kvcache. \\
\noindent\textbf{Replica-level metrics.} These include metrics such as the batch size, the number of tokens processed in each iteration, busy and idle times as well as the memory and compute utilization of each replica. \\
\noindent\textbf{Hardware metrics.} These capture cluster-wide GPU FLOPs and memory utilization. We plan to extend these to also capture the cluster's energy consumption.

\section{\syssearch}
\label{sec-syssearch}
When deploying an inference system, the system operator needs to take into account various aspects. For example, there may be SLOs on latency metrics such as TTFT and TBT or minimum QPS that needs to be supported. At the same time, the operator can try multiple configurations such as the GPU SKU (e.g. A100 vs H100) to use for deployment, the parallelization strategy (TP vs PP), scheduling policy (Orca, vLLM, Sarathi-Serve, etc.), replication degree, etc. \syssearch is a tool which helps find the optimal cost configurations to deploy an inference system while satisfying the desired SLO constraints. \syssearch leverages our simulator to compute the optimal configuration in an efficient manner. Along with the optimal configuration, \syssearch also gives detailed visualizations of how changes in configurations impact cost, TTFT, TBT, etc.

\begin{figure*}[t]
    \centering
    \begin{subfigure}[b]{\linewidth}
        \centering
        \includegraphics[width=0.9\linewidth]{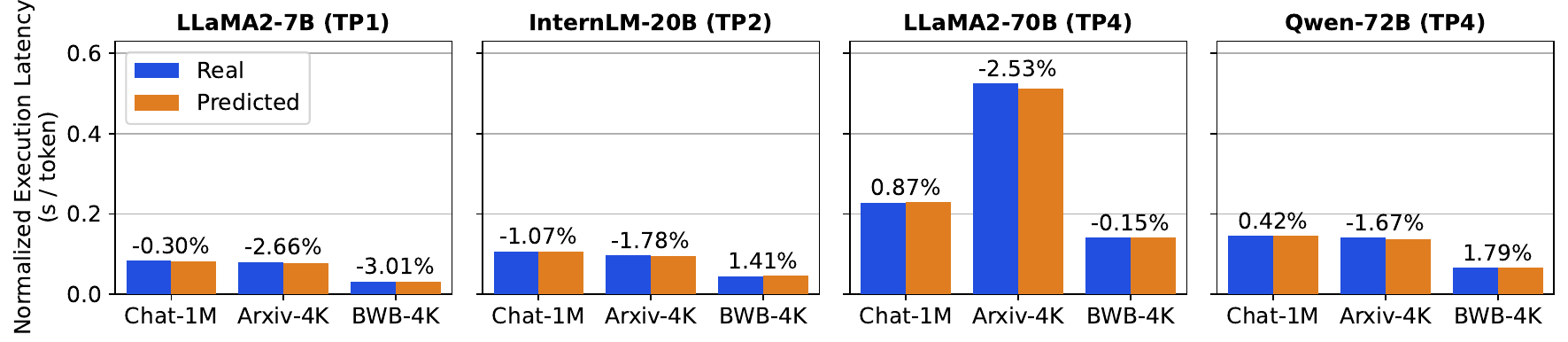}
        \caption{\small Median Normalized Execution Latency}
        \label{fig:fidelty-static-trace-p50}
    \end{subfigure}
    \begin{subfigure}[b]{\linewidth}
        \centering
        \includegraphics[trim={0 0 0 17.8}, clip, width=0.9\linewidth]{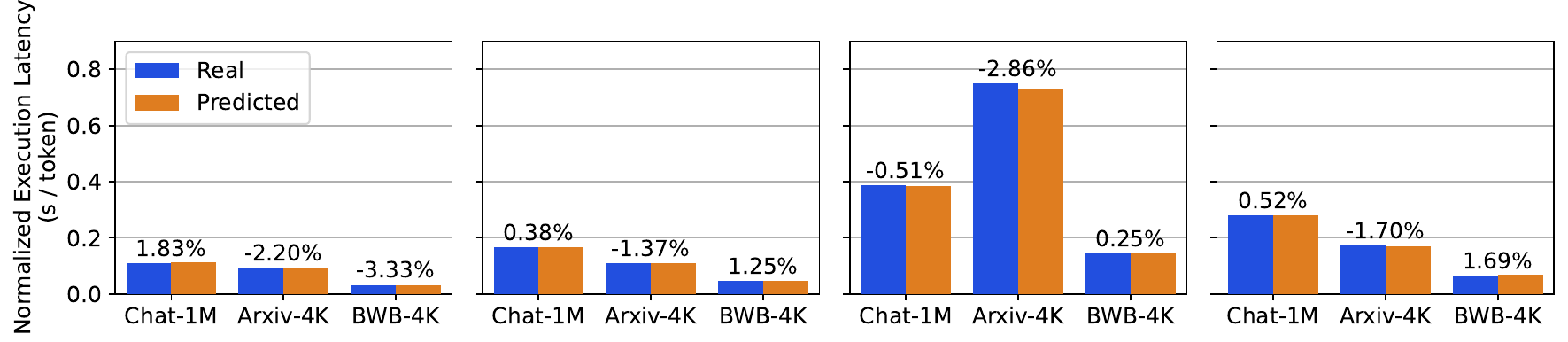}
        \caption{\small P95 Normalized Execution Latency}
        \label{fig:fidelty-static-trace-p95}
    \end{subfigure}
    \caption{Fidelity of \sysname's request execution time prediction for four models and three \textit{static} traces.}
    \label{fig:fidelity-static-trace}
\end{figure*}

\noindent\syssearch has the following main components:

\vheading{Input.} The input to the search tool consists of the LLM model, the workload (request characteristics can significantly affect inference performance), available GPU SKUs, and maximum number of GPUs in a replica.

\vheading{Constraints.} SLOs on metrics such as TTFT and TBT.

\vheading{Search space.} The search tool has the freedom to configure the parallelism strategy (TP vs PP), parallelism degree, scheduling policy, scheduler specific parameters (e.g. chunk size in Sarathi), batch size, choice of GPU, SKU, etc.

\vheading{Optimization objective.} \syssearch helps the operator maximize QPS per dollar. Consider a deployment with 16 A100 GPUs. \textit{Capacity} of the system is defined as the maximum queries per second that it can support without the queuing delay blowing up. Specifically we constrain the P99 scheduling delay to be under 5 seconds. This QPS value is divided by the cost of renting 16 A100 GPUs per hour to get the QPS per dollar value.

Given the above, \syssearch needs to solve a constrained optimization problem to find the optimal configuration in the search space. \syssearch starts with first enumerating all possible deployment configurations of the system. For each configuration, we can run our simulator on the input workload at a specified QPS and predict the metrics such as TTFT and TBT. Note, however, that the possible QPS values to pass to the simulator can be infinite. To get around this, we instead target to find the maximum QPS that a given configuration can support. We do this by tracking the scheduling delay of requests for a given configuration and QPS. Note that any system configuration will have a maximum QPS capacity for a given workload at which it can process the input requests without accumulating the request queue. We use this property to find the maximum QPS supported by a system via a simple binary search which searches for the maximum QPS which does not increase the scheduling delay beyond a threshold. Each step of this binary search involves running our simulator for the corresponding configuration and QPS. We parallelize these runs by running each search on a separate core. After this search, we have for each configuration, the maximum QPS which is supported by the system. Finally, \syssearch analyzes this data to output the optimal configuration and also generates visualizations of how changes in configurations impact the various metrics.

Since the number of configurations that need to be evaluated can be very large (in 1000s), doing a na\"ive search on actual hardware will be extremely costly. At the same time, a suboptimal choice of configuration can be very costly in the long run. Moreover, since the optimal configuration depends on the input workload, and the workload can change over time; it may be prudent to repeat this search whenever the workload characteristics have diverged from the original workload. The use of simulator in \syssearch makes this practical, by reducing this search cost by many orders of magnitude. We leverage \syssearch for our \textit{what-if} analysis in ~\autoref{sec-eval-whatif}.

Note that while \syssearch is primarily designed for configuration optimization of online serving systems, it can be repurposed for offline inference scenarios by changing the objective function from QPS per Dollar to an alternate objective like the makespan metric.

\begin{figure*}[t]
    \centering
    \begin{subfigure}[b]{\linewidth}
        \centering
        \includegraphics[width=0.9\linewidth]{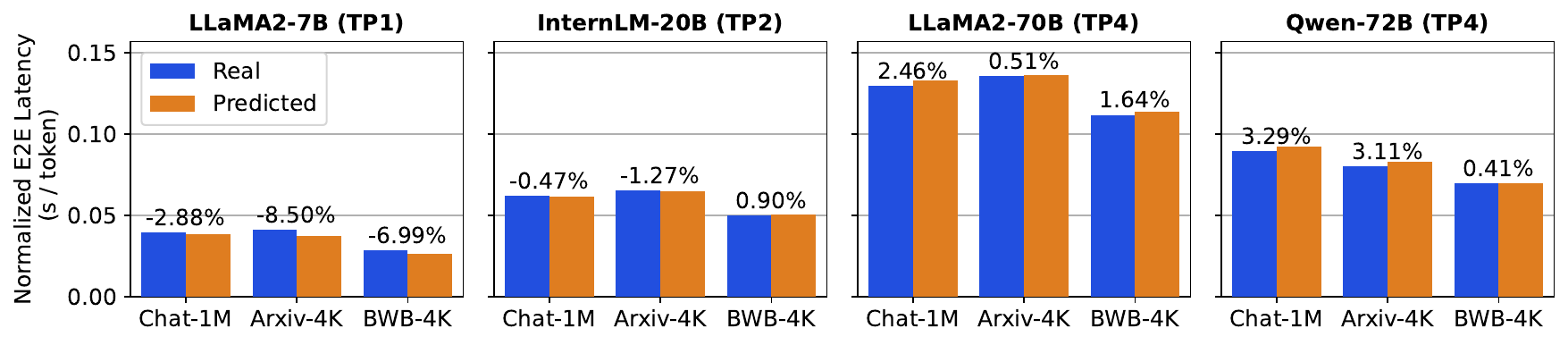}
        \caption{\small Median normalized end-to-end  latency.}
        \label{fig:fidelty-dynamic-trace-p50}
    \end{subfigure}
    \begin{subfigure}[b]{\linewidth}
        \centering
        \includegraphics[trim={0 0 0 15}, clip, width=0.9\linewidth]{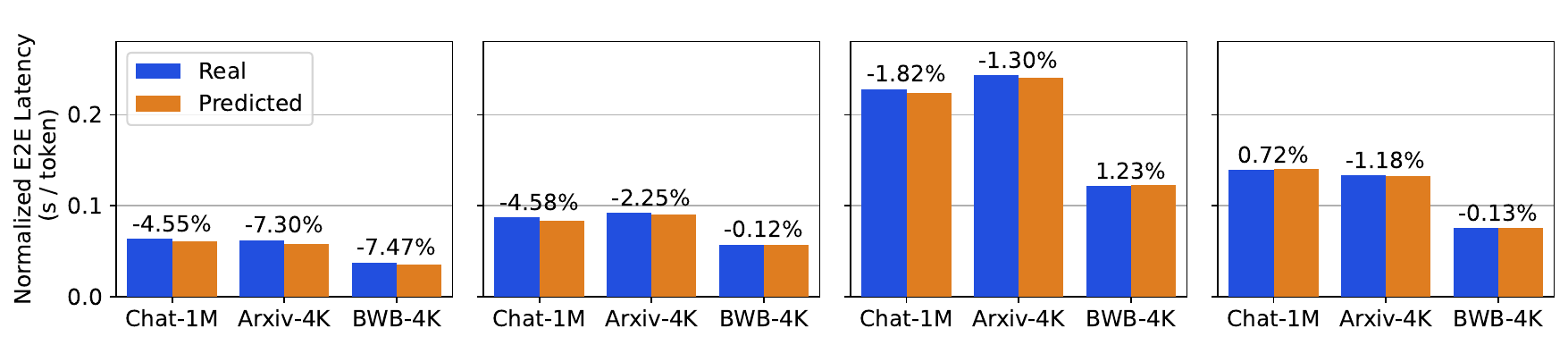}
        \caption{\small P95 normalized end-to-end  latency.}
        \label{fig:fidelty-dynamic-trace-p95}
    \end{subfigure}
    \caption{Fidelity of \sysname's execution time predictions across four models and three \textit{dynamic} workload traces, using request load at 85\% of the maximum serving capacity for each scenario.}
    \label{fig:fidelty-dynamic-trace}
\end{figure*}

\section{Evaluation}
\label{sec-eval}

In this section, we demonstrate the fidelity and usefulness of \sysname across a wide range of models, hardware configurations and workloads. We perform all our evaluations on an optimized version of the vLLM codebase, with support for different scheduling policies and CUDA graphs, which eliminates unnecessary CPU overheads. Our evaluation seeks to answer the following questions:

\vspace{-1.25em}

\begin{enumerate}
    \item Can \sysname accurately predict the end-to-end performance metrics across models of different sizes, parallelization strategies and workload traces with varying request lengths and arrival patterns (\sref{sec-eval-e2e})?
    \item Can \sysname answer what-if questions related to LLM deployment challenges for a given hardware configuration (\sref{sec-eval-whatif})?
\end{enumerate}

\vspace{-1.25em}

\subsection{Evaluation Setup}
\label{sec-eval-setup}
\vheading{Implementation.} As baseline, we use a fork of the open-source implementation of \vllm \cite{vllmpaper, vLLM:github}. %
We extend the base vLLM codebase to support various scheduling policies, chunked prefills \cite{sarathi-serve}, and an extensive telemetry system. %

\vheading{Models and Environment.} We evaluate \sysname across four models: LLaMA2 7/70B \cite{touvron2023llama}, \internlmM \cite{internlm}, and \qwenL \cite{qwen}. We use Azure \textit{Standard\_NC96ads\_A100\_v4} VMs, each equipped with 4 NVIDIA 80GB A100 GPUs, connected with pairwise NVLink. Our H100 VMs have 4 NVIDIA H100s each with 80GB memory and connected with pairwise NVLink.

\vheading{Workloads.} In order to emulate the real-world serving scenarios, we generate traces by using the request length characteristics from \chat, \arxivL and \bwbL. \chat contains one million real-world conversations with many state-of-the-art LLMs. A conversation may contain multiple rounds of interactions between the user and chatbot. Each such interaction round is performed as a separate request to the system. This multi-round nature leads to high relative variance in the prompt lengths. \arxivL is a collection of scientific publications and their summaries (abstracts) on arXiv.org \cite{arxiv}. This dataset contains large prompts and lower variance in the number of output tokens, and is representative of LLM workloads such as Microsoft M365 Copilot \cite{microsoftcopilot} and Google Duet AI \cite{googleduetai}. \bwbL is a document-level Chinese--English parallel dataset. It consists of Chinese online novels across multiple genres and their corresponding English translations. The number of output tokens outweighs the number of prompt tokens in this dataset. This dataset also has a lower variance in number of prompt and decode tokens across requests. We restrict the total request length to 4096 tokens based on the maximum context supported by the LLaMA2 family of models. We call these shortened traces \chatshort, \arxivSshort and \bwbSshort respectively. Together, these traces represent varying workload characteristics, e.g., \bwbSshort has 10\myx longer decodes and 2\myx longer prefills compared to \chatshort; and a Prefill:Decode (P:D) ratio of 0.65 compared to 2.3. Further details for these workloads are present in \autoref{table:workloads}.

\subsection{Simulator Fidelity}
\label{sec-eval-e2e}

In this section, we demonstrate \sysname's fidelity on end-to-end request-level predictions across the four models and three workloads detailed in \autoref{sec-eval-setup}. We use tensor parallel for \internlmM (TP2), \llamaL (TP4), and \qwenL (TP4). We use the default vLLM scheduler for all these experiments.
We first evaluate \sysname using static (offline) workloads where all requests are assumed to have arrived before the system starts. We then evaluate \sysname using a dynamic (online) workload in which we assume requests arrive based on a Poisson distribution, with the arrival rate corresponding to the throughput of the system.

\vheading{Evaluation Metric.} For dynamic workloads, we compare the percentage error of \sysname predictions for normalized end-to-end latency, which captures the request’s end-to-end latency divided by its output length \cite{orca, vllmpaper}. We augment this metric slightly for static workload, and measure only the request execution time, excluding the scheduling delay -- which would otherwise dominate the latency measurement. This allows us to perform more fine-grained analysis of \sysname's capability.

\vheading{Static Workloads.} We present the request latency fidelity evaluation in~\autoref{fig:fidelity-static-trace}. We observe that \sysname predicts even the tail latency (P95) with upto 3.33\% error across the four models and three datasets. Note that we observe slightly higher average error rates for the 7B model, we attribute this to the higher CPU overhead for smaller models. \amey{Explain the anomalous cases -- a) arxiv-70B - NOT DONE b) chat-P95-72B - DONE}

\vheading{Dynamic Workloads.} Next we present the evaluation of \sysname on dynamic workloads. In order to perform this evaluation, first we need to determine the request arrival rate at which we should perform this comparison. If the chosen arrival rate is too low, the system would have high idle time which is not an interesting scenario. On the other hand, if the request arrival rate is too high, the system would be overloaded where scheduling delay grows rapidly. Therefore, we evaluate \sysname's fidelity near the \textit{capacity point}, which represents the maximum arrival rate the system can sustain without overloading (\autoref{sec-syssearch}).

As shown in \autoref{fig:fidelty-dynamic-trace}, \sysname achieves high fidelity ($< 5\%$ error) in almost all scenarios with request rate set to 85\% of the system capacity -- which is reflective of real production scenarios. Note that, as we approach capacity point, any small deltas in prediction can lead to significant blow up of the errors. This is because at capacity, the system is at a tipping point -- where even slight increase in the arrival rate or request processing time leads to a sharp increase in the request latency due to uncontrolled queue delays. If either the actual or simulated system runs into overload condition, the latency numbers become hard to reconcile due to large scheduling delay. However, production systems are provisioned with a buffer so that they don't tip over the critical point due to sudden bursts. Since \sysname achieves high fidelity even at high arrival rates of up to 85\% of capacity -- making it valuable in QPS range of importance. We provide additional results at different arrival rates in \autoref{sec-appendix}.

\subsection{What-if Analysis}
\label{sec-eval-whatif}

We leverage \syssearch for an extensive \textit{what-if} analysis to understand how the performance of a configuration changes with the workload, and how the cost of serving is impacted by Service Level Objective (SLO) requirements.

\vheading{Inputs.} We find the optimal deployment configuration (one that maximizes QPS per dollar) for four models on three (dynamic) workloads described in \autoref{sec-eval-setup}. We allow choosing between the GPU SKUs of A100 and H100. The maximum number of GPUs available across replicas is set to 16. %

\begin{figure*}[t]
    \centering
    \begin{subfigure}[b]{\linewidth}
        \centering
        \includegraphics[width=0.9\linewidth]{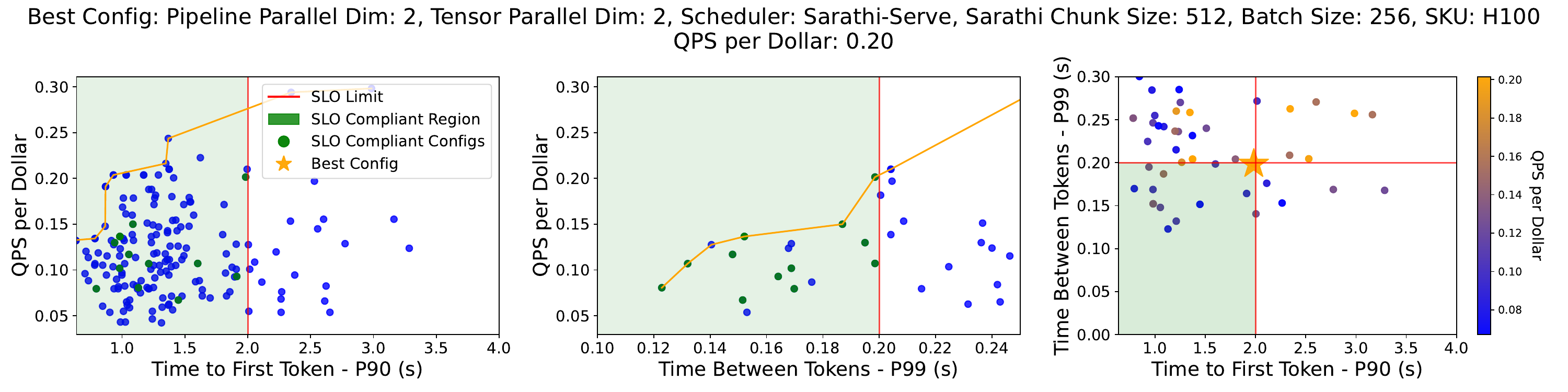}
        \caption{\small \llamaL -- \chat}
        \label{fig:pareto-slo-chat}
    \end{subfigure}
    \hfill
    \begin{subfigure}[b]{0.9\linewidth}
        \centering
        \includegraphics[width=\linewidth]{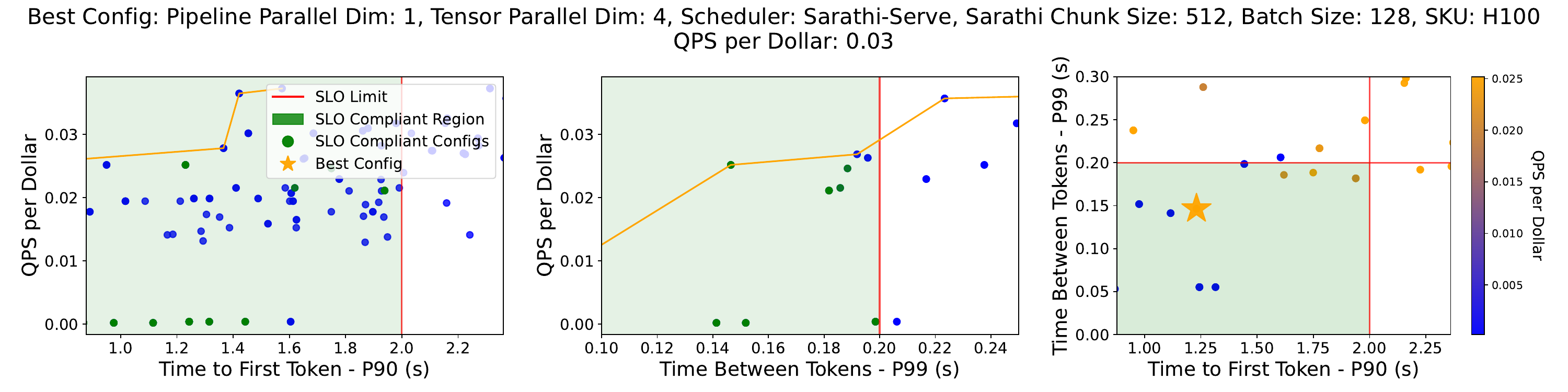}
        \caption{\small \qwenL -- \arxivSshort}
        \label{fig:pareto-slo-bwb}
    \end{subfigure}
    \caption{
   Capacity per dollar for different deployment configurations vs corresponding TTFT-P90 (left) and  TBT-P99 (middle). Also show is the Pareto curve for these configurations. Shaded area corresponds to region where the corresponding SLO is satisfied. (right) Both latency metrics for these configuration, with capacity per dollar visualized via a temperature colormap. In the left and middle plots, green points correspond to configurations which satisfy SLOs for both metrics. Note that blue points on a Pareto curve show that, even Pareto curve points for one metric may not satisfy SLO for the other metric.
   }
    \label{fig:pareto-curves}
\end{figure*}

\vheading{SLOs.} We put the following SLO constraints on the latency metrics: TTFT P90 $<$ 2s and TBT P99 $<$ 200ms. We use a more relaxed constraint of P90 for TTFT since it is a one time delay experienced by the user, as opposed to TBT which is recurrent for each output token.

\vheading{Deployment Configurations.} We experiment with TP and PP dimensions of 1, 2 and 4 for each, with three iteration-level schedulers \vllm, \orcaplus and \sarathi that dynamically allocate memory for \kvcache using paged attention. \vllm is a throughput-oriented scheduler that maximizes batch size by eagerly scheduling prefills while pausing on-going decodes. \orcaplus is Orca~\cite{orca} implemented over \vllm's paged attention. \sarathi creates hybrid batches with partial prefills to avoid pausing decodes while keeping GPU utilization high. We try these schedulers with batch size 32, 64, 128, 256 and 512. Note that the batch size gets divided by number of microbatches with PP. \vllm and \orcaplus have a limit of maximum 4096 tokens per iteration while \sarathi has max 512, 1K and 2K tokens per iteration (also known as chunk size).

\begin{figure}[t]
    \centering
    \hspace{-2em}
    \includegraphics[width=0.45\textwidth]{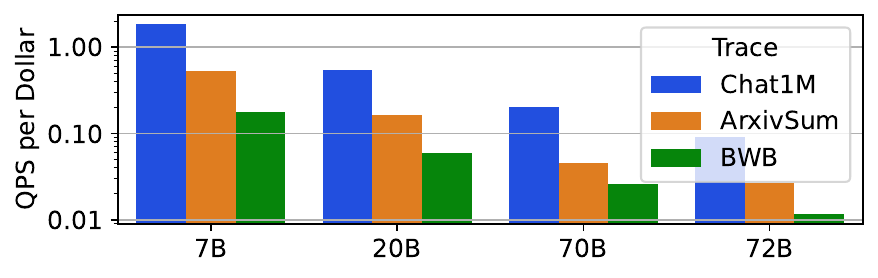}
    \caption{\small
        QPS per dollar for best configurations using P90 TTFT and P99 TBT SLOs of 2s and 200ms respectively.}
    \label{fig-cap-per-dol}
\end{figure}

\autoref{fig:intro:optimal-configs} shows the optimal configuration for the three models for each of the workloads, and ~\autoref{fig-cap-per-dol} shows the QPS per dollar for the optimal configuration. We summarize the key takeaways below.

First, the \textit{change in workload can drastically change the optimal configuration}. For example, for the LLama2-70B model, the optimal configuration for LMSys-Chat-1M uses batch size of 256, while for BWB it is 64. This is a consequence of the high \kvcache load in BWB workload due to large decode sequences. Even the optimal GPU SKU changes from H100 for Chat-1M to A100 for BWB. 

Second, even models with similar sizes can have very different performance characteristics due to variation in architectural details. For instance, \llamaL uses Group Query Attention (GQA), where as \qwenL employs Multi Head Attention (MHA) -- which translates to 8\myx higher \kvcache load. As a result, \qwenL is almost 2\myx more costly to serve and requires a different deployment configuration.

Finally, from \autoref{fig-cap-per-dol} it is clear that the capacity per dollar follows the expected trend. For example, larger models have lower capacity compared to smaller models. Also, Chat-1M has the least cost due to fewer prefill and decode tokens, while BWB has the highest cost due to larger number of tokens, especially decode tokens which are more expensive to compute compared to prefill. This complete exploration costs only 125 US dollars in simulation as opposed to actual execution which would have required 1.14 million dollars. We provide a detailed cost comparison in~\autoref{table:tbl-eval-whatif}.

\vheading{Configuration Stability.} \autoref{fig:intro:cost-of-misconfig} shows the overhead factor of using the optimal configuration for one workload, to serve a different workload on the \llamaL model. As shown, such a misconfiguration can result in a very high overhead, e.g., running \chat workload with the optimal configuration of \arxivS workload results in a 2\myx overhead! This shows that even for the same model, the cost of using a homogeneous deployment configuration can result in huge overheads, as the optimal configuration for one workload can be far from optimal for another workload.

\vheading{Pareto Frontier Analysis.}  We next analyze the Pareto frontier produced by \sysname for \llamaL-\chat and \qwenL-\bwbS workloads.~\autoref{fig:pareto-curves} shows the best QPS per dollar for different configurations and the corresponding TTFT-P90 (left), TBT-P99 metrics (middle) along with the SLO complient regions. The figures on the right plot both the latency metrics for these configuration, and visualize the QPS per dollar via a temperature colormap. We summarize the key takeaways.

First, \textit{configurations which are optimal on
one metric may not satisfy the SLO constraint on the
other metric} (these are the blue points on the Pareto curve). Second, \textit{small changes in latency SLOs can result in a significant cost overhead}. For example, for the \llamaL-\chat workload, if the TBT SLO is changed from 0.12 seconds to 0.14 seconds (a difference of only 20ms), the Pareto curve point moves from approximately $0.07$ to $0.13$, $\sim 1.85\times$ reduction in cost!

\section{Related Work}
\label{sec-related}

Prior techniques leverage the predictability of DNN training iterations~\cite{astra, gandiva} to model the performance of the entire job. For example, Habitat~\cite{habitat} models the performance of a training job on different types of GPUs based on the runtime profile collected of a few training iterations on a given GPU. In doing so, Habitat applies the roofline model~\cite{roofline} to estimate the performance of individual operators based on the compute and memory requirements of the operator along with the compute and memory bandwidth of a GPU. Daydream~\cite{daydream} proposes a different approach focused on modeling the effect of various system optimizations on training performance across various deployment scenarios. Daydream can help answer questions like: what is the main performance bottleneck in my training job (e.g., memory or network bandwidth), how will optimizations like kernel-fusion, quantization or gradient compression help improve performance etc.  To accurately model the effect of such optimizations, Daydream first constructs a computation graph of a training job and then applies optimizations via graph transformations (e.g., kernel-fusion can be applied by substituting individual kernel nodes with a single node that represents the fused kernels in the computation graph). Proteus~\cite{proteus} further enables simulating various parallelization strategies to identify the best partitioning and scheduling strategy for a given training job. It does so by first modeling a parallelization strategy with a unified representation called \textit{Strategy Tree} and then compiling it into a distributed execution graph. In another approach ~\cite{recommender_modeling}, the authors propose a critical-path based strategy to predict the per-batch training time of deep learning recommendation models. Different from these training-based simulators, \sysname is the first simulator that accounts for the specific properties of LLM inference.

\vspace{-0.5em}
\section{Conclusion}
\label{sec-conc}

LLM inference efficiency depends on a large number of configuration knobs such as the type or degree of parallelism, scheduling strategy, GPU SKUs. It is impractical to run all possible configurations on actual hardware. In this paper, we present \sysname: a high fidelity and easily extensible simulator for LLM inference, along with a benchmark and search suite. \sysname answers deployment related what-if questions that identify efficient deployment strategies for production environments and helps in evaluating the efficacy of various systems optimizations at nominal cost.

\bibliography{all}
\bibliographystyle{mlsys2024}

\appendix
\newpage
\columnbreak
\columnbreak
\clearpage

\section{Appendix}
\label{sec-appendix}

\subsection{Impact of Request Arrival Rate on Fidelity for Dynamic Workloads}

\begin{figure*}
    \centering
        \begin{subfigure}[b]{\linewidth}
        \centering
        \includegraphics[width=0.9\linewidth]{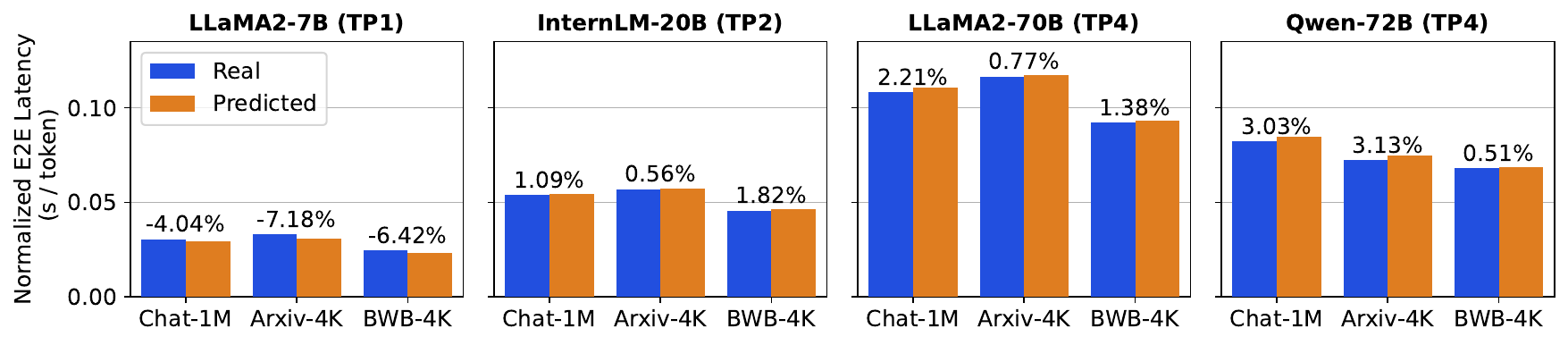}
        \caption{\small Median normalized end-to-end  latency at 75\% of Maximum Capacity}
        \label{fig:fidelty-dynamic-trace-75-p50}
    \end{subfigure}
    \begin{subfigure}[b]{\linewidth}
        \centering
        \includegraphics[width=0.9\linewidth]{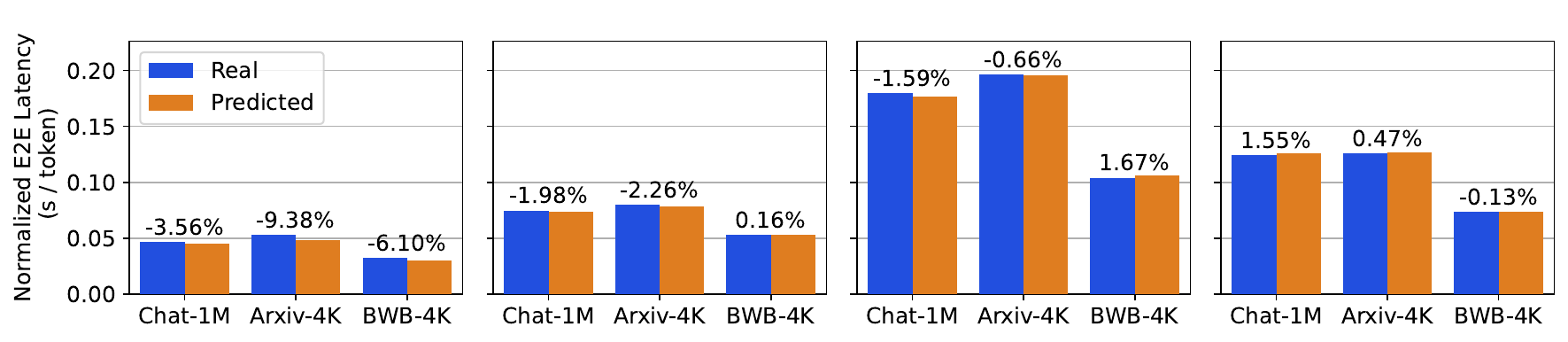}
        \caption{\small P95 normalized end-to-end  latency at 75\% of maximum capacity}
        \label{fig:fidelty-dynamic-trace-75-p95}
    \end{subfigure}
    \begin{subfigure}[b]{\linewidth}
        \centering
        \includegraphics[trim={0 0 0 17.8}, clip, width=0.9\linewidth]{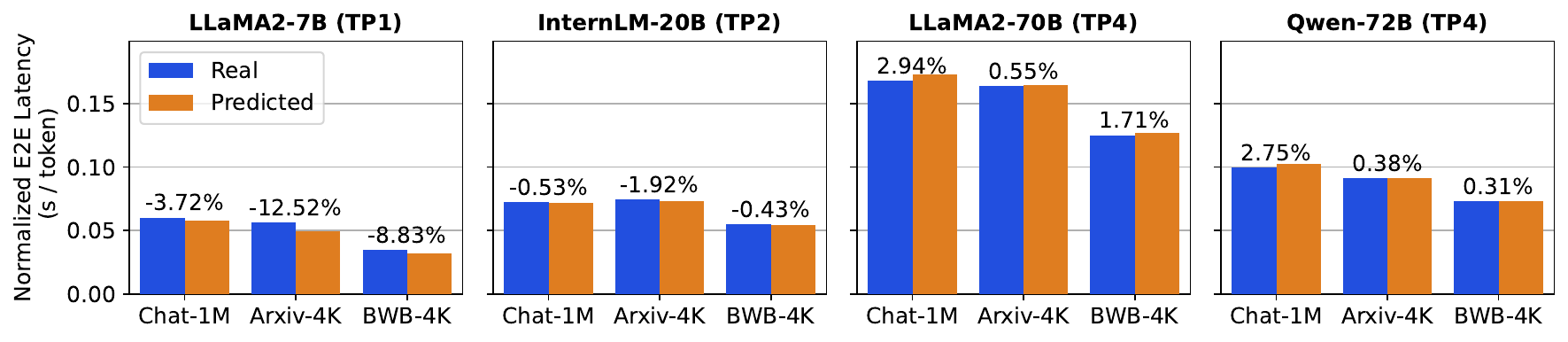}
        \caption{\small Median normalized end-to-end  latency at 95\% of maximum capacity}
        \label{fig:fidelty-dynamic-trace-95-p50}
    \end{subfigure}
    \begin{subfigure}[b]{\linewidth}
        \centering
        \includegraphics[width=0.9\linewidth]{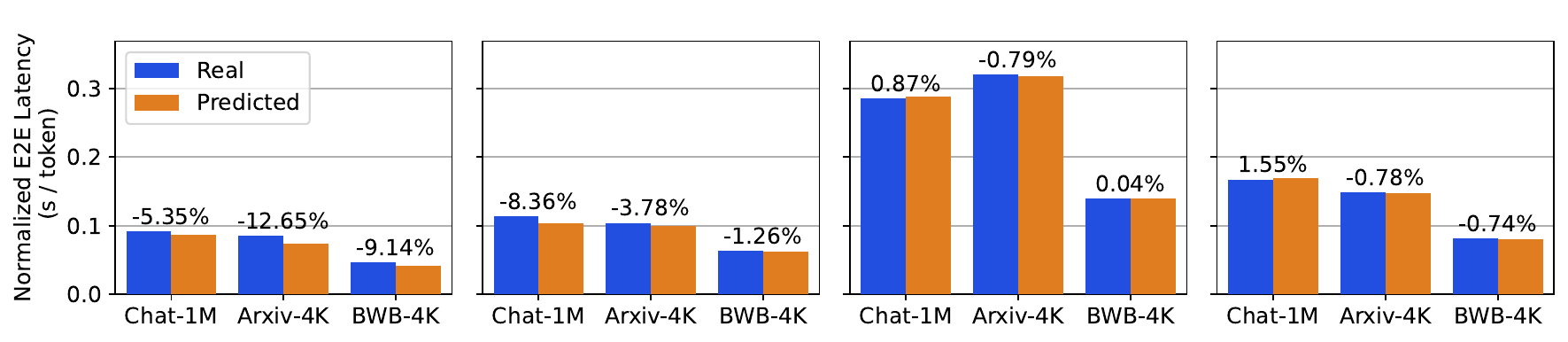}
        \caption{\small P95 normalized end-to-end  latency at 95\% of Maximum Capacity}
        \label{fig:fidelty-dynamic-trace-95-p95}
    \end{subfigure}
    \caption{\small Fidelity of \sysname's execution time predictions across four models and three \textit{dynamic} workload traces at different arrival rates.}
    \label{fig:fidelty-dynamic-trace-95}
\end{figure*}

\begin{figure*}
    \centering
        \includegraphics[width=0.8\linewidth]{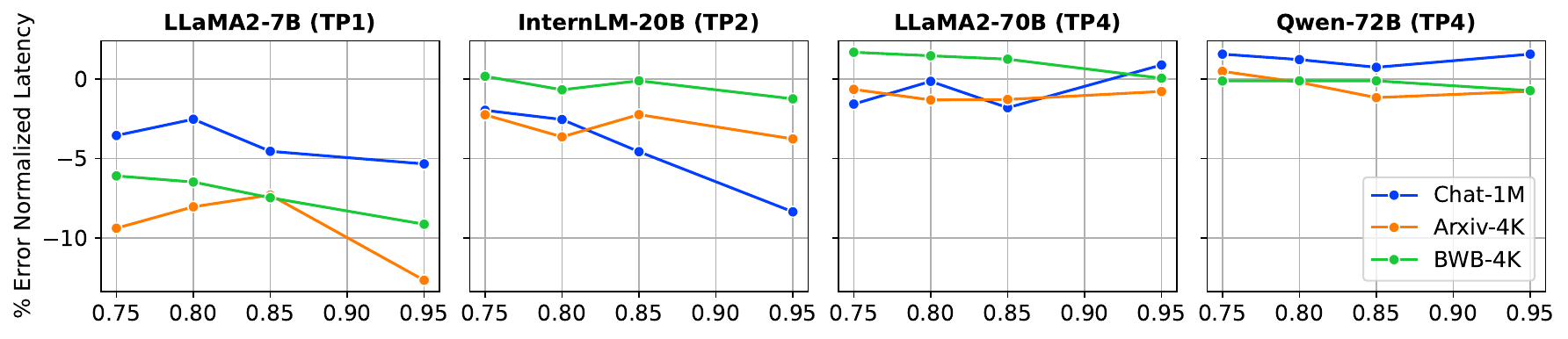}
    \caption{Prediction error for p95 normalized end-to-end latency at arrival rates between 0.75\myx and 0.95\myx of the maximum serving capacity.}
    \label{fig:fidelty-dynamic-trace-trends}
\end{figure*}

We present additional fidelity results for \sysname at different request arrival rates in \autoref{fig:fidelty-dynamic-trace-95}. We find that \sysname retains high fidelity even at 95\% of maximum system capacity for larger models, however, for \llamaS, where we have slightly higher error due to CPU overheads, the errors cascade and we see up to 12.65\% maximum error. We also provide  the error trends in \autoref{fig:fidelty-dynamic-trace-trends}.

\subsection{Cost Breakdown of What-if Analysis}

\sysname's ability to efficiently and accurate simulated complex deployment scenarios allows us to explore search configuration spaces at cheap negligible cost. The what-if analysis presented in \autoref{fig:intro:optimal-configs}, required a total of 35,565 runs, with a total projected GPU duration of 1,139,865 dollars. The same search completes takes only $\sim12.5$ hours on a 96-core CPU machine costing just \$125. The breakdown of each task is presented in \autoref{table:tbl-eval-whatif}.

\begin{table}[h]
\centering
\scalebox{0.9}{
\begin{tabular}{c|cc|ccc} \hline
\multirow{2}{*}{Scenario} & \multicolumn{2}{|c|}{Time} & \multicolumn{3}{|c}{Cost(\$)} \\
&  Act &  Sim  & Act  & Sim & Savings  \\
\hline
7B-Chat1M & 4K hrs & 31 min & 20K & 5 & 3837x \\
7B-Arxiv & 10K hrs & 47 min & 52K & 8 & 6708x \\
7B-BWB & 18K hrs & 136 min & 97K & 22 & 4324x \\
20B-Chat1M & 6K hrs & 21 min & 33K & 3 & 9518x \\
20B-Arxiv & 14K hrs & 25 min & 73K & 4 & 17746x \\
20B-BWB & 16K hrs & 52 min & 84K & 9 & 9805x \\
70B-Chat1M & 12K hrs & 21 min & 64K & 4 & 18151x \\
70B-Arxiv & 15K hrs & 16 min & 78K & 3 & 30187x \\
70B-BWB & 15K hrs & 27 min & 77K & 4 & 17333x \\
72B-Chat1M & 14K hrs & 43 min & 76K & 7 & 10726x \\
72B-Arxiv & 17K hrs & 16 min & 88K & 3 & 33354x \\
72B-BWB & 18K hrs & 25 min & 93K & 4 & 22102x \\
\bottomrule
\end{tabular}}
\caption{Cost of finding the optimal deployment configuration.}
\label{table:tbl-eval-whatif}
\end{table}

\end{document}